# Response to Comment on "All-optical machine learning using diffractive deep neural networks"

*Deniz Mengu*[1,2,3], *Yi Luo*[1,2,3], *Yair Rivenson*[1,2,3], *Xing Lin*[1,2,3], *Muhammed Veli*[1,2,3], *Aydogan Ozcan*[1,2,3,4,*]

[1] Electrical and Computer Engineering Department, University of California, Los Angeles, CA, 90095, USA
[2] Bioengineering Department, University of California, Los Angeles, CA, 90095, USA
[3] California NanoSystems Institute, University of California, Los Angeles, CA, 90095, USA
[4] Department of Surgery, David Geffen School of Medicine, University of California, Los Angeles, CA, 90095, USA.

†Equal contributing authors.

[*]Email: ozcan@ucla.edu

Keywords: Optical computing, Deep networks, Deep learning, Optical neural networks, Diffractive Optical Networks, $D^2NN$

## ABSTRACT

In their Comment, Wei *et al.* (arXiv:1809.08360v1 [cs.LG]) claim that our original interpretation of Diffractive Deep Neural Networks ($D^2NN$) represent a 'mischaracterization' of the system due to linearity and passivity. In this Response, we detail how this 'mischaracterization' claim is unwarranted and oblivious to several sections detailed in our original manuscript (*Science*, DOI: 10.1126/science.aat8084) that specifically introduced and discussed optical nonlinearities and reconfigurability of $D^2NNs$, as part of our proposed framework to enhance its performance. To further refute the 'mischaracterization' claim of Wei *et al.*, we, once again, demonstrate the "depth" feature of optical $D^2NNs$ by showing that multiple diffractive layers operating collectively within a $D^2NN$ present additional degrees-of-freedom compared to a single diffractive layer to achieve better classification accuracy, as well as improved output signal contrast and diffraction efficiency as the number of diffractive layers increase, showing the deepness of a $D^2NN$, and its inherent depth advantage for improved performance. In summary, the Comment by Wei *et al.* does not provide an amendment to the original teachings of our original manuscript, and all of our results, core conclusions and methodology of research reported in *Science* (DOI: 10.1126/science.aat8084) remain entirely valid.

## 1- Introduction to Diffractive Deep Neural Networks ($D^2NN$)

We have recently introduced an optical machine learning framework, termed as Diffractive Deep Neural Network ($D^2NN$) (*1*), where deep learning and error back-propagation methods are used to design, using a computer, diffractive layers that collectively perform a desired task that the network is trained for. In the training phase of a $D^2NN$, the transmission and/or reflection coefficients of the individual pixels (i.e., neurons) of each layer are optimized such that as the light diffracts from the input plane toward the output plane, it computes the task at hand. Once this training phase in a computer is complete, these layers are physically fabricated and stacked together to form an all-optical network that executes the trained function without the use of any power, except for the illumination light and the output detectors.

In our previous work, we experimentally demonstrated the success of $D^2NN$ framework at THz part of the electromagnetic spectrum and used a standard 3D-printer to fabricate and assemble together the designed $D^2NN$ layers (*1*). In addition to demonstrating optical classifiers, we also demonstrated that the same $D^2NN$ framework can be used to design an imaging system by 3D-engineering of optical components using deep learning (*1*). In these earlier results, we used coherent illumination and encoded the input information in phase or amplitude channels of different $D^2NN$ systems. Another important feature of $D^2NN$s is that the axial spacing between the diffractive layers is very small, e.g., less than 50 wavelengths ($\lambda$), which makes the entire design highly compact and flat (*1*). A recent work (*2*) has expanded our analysis on this $D^2NN$ framework, further improving its all-optical inference and classification performance, also showing the integration of $D^2NN$s with electronic neural networks to create hybrid classifiers that significantly reduce the number of input pixels into an electronic network using an ultra-compact front-end $D^2NN$ with a layer-to-layer distance of a few wavelengths.

Our initial experimental demonstration of $D^2NN$s was based on linear materials, without including the equivalent of a nonlinear activation function within the optical network, as stated in our original manuscript (*1*). As detailed in Ref. (*1*), optical nonlinearities can also be incorporated into a $D^2NN$ using non-linear materials including e.g., crystals, polymers or semiconductors, to further improve its inference performance using nonlinear optical effects within diffractive layers. For such a nonlinear $D^2NN$ design, resonant nonlinear structures (based on e.g., plasmonics or metamaterials, see (*1*)) tuned to the illumination wavelength would be important to lower the required intensity levels.

It is important to emphasize that, even using linear optical materials to create a $D^2NN$, the optical network shows "*depth*", i.e., a single diffractive layer does not possess the same degrees-of-freedom to achieve the same level of *classification accuracy*, *signal contrast* and *power efficiency* at the output plane that multiple diffractive layers can collectively achieve for a given task. This is unfortunately entirely missed by Wei *et al.* in their Comment (*3*). It is true that, for a linear diffractive optical network, the entire wave propagation and diffraction phenomena that happen between the input and output planes can be squeezed into a single matrix operation; *however*, this arbitrary mathematical operation that is defined by multiple learnable diffractive layers cannot be performed in general by a single diffractive layer placed between the same input and output planes, and therefore a general multilayer $D^2NN$ designed by deep learning is not *optically* squeezable into a single diffractive layer in terms of its classification or inference performance. That is why, multiple diffractive layers forming a $D^2NN$ show the *depth* advantage and statistically perform much better compared to a single diffractive layer trained for the same classification task, as also reported in the supplementary materials of Ref. (*1*), and in Ref. *(2)*.

Starting with the next sub-section we will expand on some of these points, refuting the 'mischaracterization' claims of Wei *et al*. ***In conclusion, all the results, the core conclusions and the methodology of research in our original manuscript*** (*1*) ***remain entirely valid***.

*2- $D^2NN$ framework exhibits depth*

In addition to the supplementary materials of Ref. (*1*), the depth feature of $D^2$NNs has been further investigated and quantified in Ref. (*2*). As illustrated in Figure 1 below, multiple diffractive layers operating collectively within a $D^2$NN design present additional degrees-of-freedom compared to a single diffractive layer to achieve better classification accuracy, as well as improved diffraction efficiency and signal contrast at the output plane of the network; the latter two are especially important for experimental implementations of all-optical diffractive networks as they dictate the required illumination power levels as well as signal-to-noise ratio related error rates for all-optical classification tasks involving $D^2$NNs.

Stated differently, $D^2$NN framework, even when it is composed of linear optical materials, is "Deep" and shows "Depth" advantage because an increase in the number of diffractive layers (1) improves its statistical inference accuracy (see Figs. 1A and 1D, for MNIST and Fashion MNIST datasets, respectively), and (2) improves its overall power efficiency and the signal contrast at the correct output detector with respect to the detectors assigned to other classes (see Figs. 1B,C and Figs. 1E,F for MNIST and Fashion MNIST datasets, respectively). Here, we defined the *power efficiency* of a $D^2$NN as the percentage of the optical signal detected at the target label detector corresponding to the correct data class, with respect to the total optical signal at the output plane of the optical network. Furthermore, we defined the *output signal contrast* as the difference between the optical signal captured by the target label detector corresponding to the correct data class and the maximum signal detected by the rest of the detectors (i.e., the strongest competitor detector for each test sample), normalized with respect to the total optical signal at the output plane. Therefore, for a given, available input illumination power and detector signal-to-noise ratio, the overall error rate of the all-optical network decreases as the number of diffractive layers increase. All these highlight the *depth* feature of a $D^2$NN and the associated advantages.

This is ***not*** in contradiction with the fact that, for an all-optical $D^2$NN that is made of linear optical materials, the entire diffraction phenomenon that happens between the input and output planes can be squeezed into a single matrix operation (in reality, every material exhibits some volumetric and surface nonlinearities, and what we mean here by a linear optical material is that these effects are negligible). In fact, an arbitrary mathematical operation that is defined by multiple learnable diffractive layers cannot be performed in general by a single diffractive layer placed between the same input and output planes; additional optical components and/or layers would be needed to all-optically perform an arbitrary mathematical operation that multiple learnable diffractive layers can in general perform.

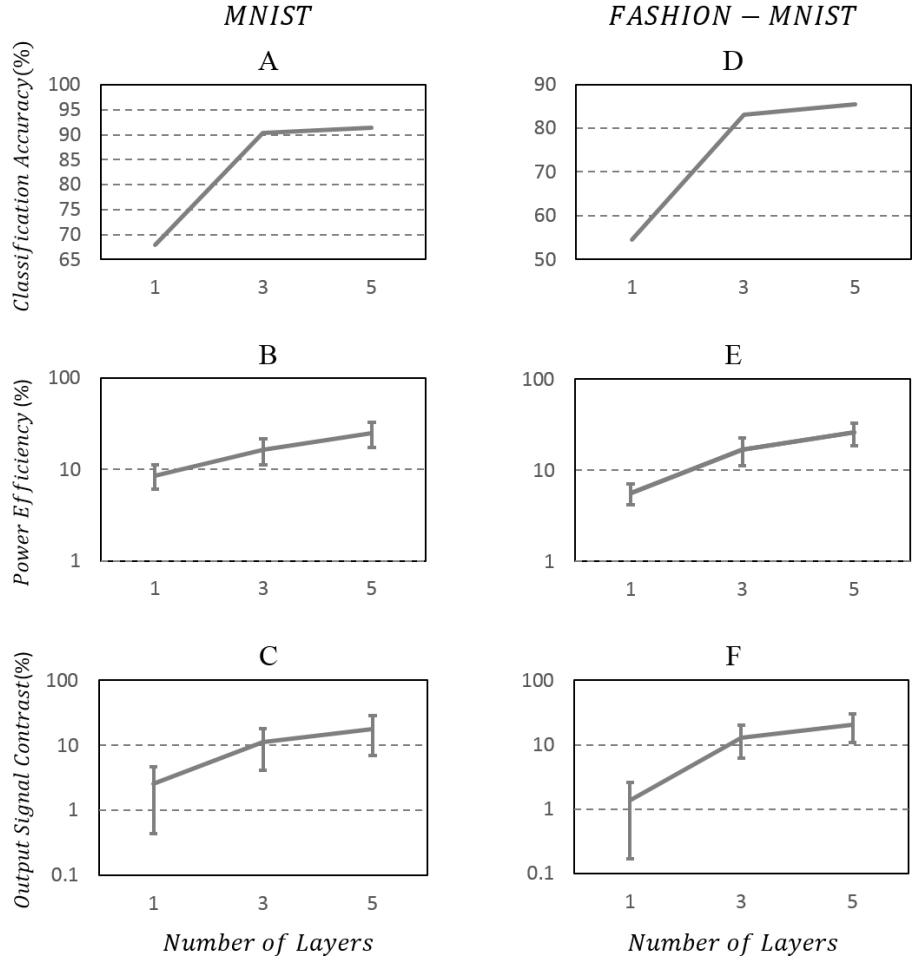

**Fig. 1: Classification accuracy, power efficiency and output signal contrast for phase-only D²NN classifier designs with 1, 3 and 5-layers.** *(A)* Blind testing accuracy, *(B)* power efficiency and *(C)* output signal contrast analysis of phase-only all-optical classifiers trained for handwritten digits (MNIST dataset). *(D-F)* are the same as *(A-C)*, only the classified dataset is Fashion-MNIST instead. We define the power efficiency of a D²NN as the percentage of the optical signal detected at the target label detector corresponding to the correct data class, with respect to the total optical signal at the output plane of the optical network.*(2)* We define the output signal contrast as the difference between the optical signal captured by the target label detector corresponding to the correct data class and the maximum signal detected by the rest of the detectors (i.e., the strongest competitor detector for each test sample), normalized with respect to the total optical signal at the output plane.*(2)* MSE loss function was used in this analysis, and the same conclusion applies for other loss functions as well. For further details on this analysis, refer to Ref. *(2)*.

Our D²NN framework creates a unique opportunity to use deep learning to design multiple diffractive layers, within a very tight layer-to-layer spacing of less than 50×λ, that collectively function as an all-optical classifier, and this framework will further benefit from nonlinear optical materials and resonant optical structures based on e.g., plasmonics or metamaterials to further enhance its inference performance. These have *already* been emphasized in our original manuscript, as we will detail in the next sub-section with "quotes" from Ref. *(1)*.

In addition to the above highlighted *depth/deepness* feature of a D²NN, another strong connection between deep learning and the D²NN framework is the fact that standard deep learning tools available in e.g., TensorFlow are at the heart of engineering of D²NN layers to all-optically execute machine learning tasks, which forms the basis of $D^2NNs$ and one of the core teachings of Ref. (*1*).

## *3- Linear vs. Nonlinear Optical Materials in $D^2NNs$*

As emphasized earlier, our experimental demonstration of $D^2NNs$ was based on linear materials, without including the equivalent of a nonlinear activation function within the optical network as stated in our original manuscript (*1*). In Ref. (*1*) we provided detailed discussions on optical nonlinearity in $D^2NN$ framework, and the Comment of Wei *et al.* reflects ignorance of several related parts of our manuscript (listed and quoted below), which forms one of the sources of the unjustified 'mischaracterization' claim of their Comment.

• Supplementary Materials,(*1*) Sub-section "**Comparison with standard deep neural networks**"

"…Compared to standard deep neural networks, a $D^2NN$ is not only different in that it is a physical and all-optical deep network, but also it possesses some unique architectural differences. First, the inputs for neurons are complex-valued, determined by wave interference and a multiplicative bias, i.e., the transmission/reflection coefficient [also reported clearly in Figure 1 of the main text]. …. **Second, the individual function of a neuron is the phase and amplitude modulation of its input to output a secondary wave, unlike e.g., a sigmoid, a rectified linear unit (ReLU) or other nonlinear neuron functions used in modern deep neural networks. Although not implemented here, optical nonlinearity can also be incorporated into a diffractive neural network in various ways; see the sub-section "Optical Nonlinearity in Diffractive Neural Networks"**…". (*1*)

• Supplementary Materials,(*1*) Sub-section "**Optical Nonlinearity in Diffractive Deep Neural Networks**"

"…Optical nonlinearity can be incorporated into our deep optical network design using various optical non-linear materials (crystals, polymers, semiconductor materials, doped glasses, among others as detailed below). A $D^2NN$ is based on controlling the diffraction of light through complex-valued diffractive elements to perform a desired/trained task. Augmenting nonlinear optical components is both practical and synergetic to our $D^2NN$ framework.
    Assuming that the input object, together with the $D^2NN$ diffractive layers, create a spatially varying complex field amplitude E(x,y) at a given network layer, then the use of a nonlinear medium (e.g., optical Kerr effect based on third-order optical nonlinearity, $\chi^{(3)}$) will introduce an all-optical refractive index change which is a function of the input field's intensity, $\Delta n \propto \chi^{(3)} E^2$. This intensity dependent refractive index modulation and its impact on the phase and amplitude of the resulting waves through the diffractive network can be numerically modeled and therefore is straightforward to incorporate as part of our network training phase. Any third-order nonlinear material with a strong $\chi^{(3)}$ could be used to form our nonlinear diffractive layers: glasses (e.g., $As_2S_3$, metal nanoparticle doped glasses), polymers (e.g., polydiacetylenes), organic films, semiconductors (e.g., GaAs, Si, CdS), graphene, among others. There are different fabrication methods that can be employed to structure each nonlinear layer of a diffractive neural network using these materials.
    In addition to third-order all-optical nonlinearity, another method to introduce nonlinearity into a $D^2NN$ design is to use saturable absorbers that can be based on materials such as semiconductors, quantum-dot films, carbon nanotubes or even graphene films. There are also various fabrication methods, including standard photo-lithography, that one can employ to structure such materials as part of a $D^2NN$ design; for example, in THz wavelengths, recent research has demonstrated inkjet printing of graphene saturable absorbers (*39*). Graphene-based saturable absorbers are further advantageous since they work well even at relatively low modulation intensities (*40*).
    Another promising avenue to bring non-linear optical properties into $D^2NN$ designs is to use nonlinear metamaterials. These materials have the potential to be integrated with diffractive networks owing to their compactness and the fact that they can be manufactured with standard fabrication processes. While a significant part of the previous work in the field has focused on second and third harmonic generation, recent studies have demonstrated very strong optical Kerr effect for different parts of the electromagnetic spectrum (*41-42*), which can be incorporated into our deep diffractive neural network architecture to bring all-optical nonlinearity into its operation.
    Finally, one can also use the DC electro-optic effect to introduce optical nonlinearity into the layers of a

$D^2NN$ although this would deviate from all-optical operation of the device and require a DC electric-field for each layer of the diffractive neural network. This electric-field can be externally applied to each layer of a $D^2NN$; alternatively one can also use poled materials with very strong built-in electric fields as part of the material (e.g., poled crystals or glasses). The latter will still be all-optical in its operation, without the need for an external DC field.

To summarize, there are several practical approaches that can be integrated with diffractive neural networks to bring physical all-optical nonlinearity to $D^2NN$ designs…" (*1*)

- Main text, conclusion part: "…To achieve these new technologies, nonlinear optical materials (14) and a monolithic $D^2NN$ design that combines all layers of the network as part of a 3D-fabrication method would be desirable." (*1*)

## *4- Passive vs. reconfigurable designs in $D^2NN$ framework*

The Comment of Wei *et al.* is also appallingly ignorant of an entire sub-section in our Supplementary Materials (*1*) titled "**Reconfigurable $D^2NN$ Designs**", which we quote below to emphasize how *passive* vs. *reconfigurable designs* in $D^2NN$ framework are discussed in our original manuscript, (*1*) also introducing hybrid combinations as a viable option, i.e., passive and reconfigurable diffractive layers working together to improve $D^2NN$ performance. This unfortunate obliviousness of their Comment *(3)* is another source of their unwarranted 'mischaracterization' claim, with a baseless statement on "….*overlooking… strict passivity*".

From (*1*): "**Reconfigurable $D^2NN$ Designs.** One important avenue to consider is the use of spatial light modulators (SLMs) as part of a diffractive neural network. This approach of using SLMs in $D^2NN$s has several advantages, at the cost of an increased complexity due to deviation from an entirely passive optical network to a reconfigurable electro-optic one. First, a $D^2NN$ that employs one or more SLMs can be used to learn and implement various tasks because of its reconfigurable architecture. Second, this reconfigurability of the physical network can be used to mitigate alignment errors or other imperfections in the optical system of the network. Furthermore, as the optical network statistically fails, e.g., a misclassification or an error in its output is detected, it can mend itself through a transfer learning based re-training with appropriate penalties attached to some of the discovered errors of the network as it is being used. For building a $D^2NN$ that contains SLMs, both reflection and transmission based modulator devices can be used to create an optical network that is either entirely composed of SLMs or a hybrid one, i.e., employing some SLMs in combination with fabricated (i.e., passive) layers.

In addition to the possibility of using SLMs as part of a reconfigurable $D^2NN$, another option to consider is to use a given 3D-printed or fabricated $D^2NN$ design as a fixed input block of a new diffractive network where we train only the additional layers that we plan to fabricate. Assume for example that a 5-layer $D^2NN$ has been printed/fabricated for a certain inference task. As its prediction performance degrades or slightly changes, due to e.g., a change in the input data, etc., we can train a few new layers to be physically added/patched to the existing printed/fabricated network to improve its inference performance. In some cases, we can even peel off (i.e., discard) some of the existing layers of the printed network and assume the remaining fabricated layers as a fixed (i.e., non-learnable) input block to a new network where the new layers to be added/patched are trained for an improved inference task (coming from the entire diffractive network: old layers and new layers).

Intuitively, we can think of each $D^2NN$ as a "Lego" piece (with several layers following each other); we can either add a new layer (or layers) on top of existing (i.e., already fabricated) ones, or peel off some layers and replace them with the new trained diffractive blocks. This provides a unique physical implementation (like blocks of Lego) for transfer learning or mending the performance of a printed/fabricated $D^2NN$ design.

We implemented this concept of Lego design for our Fashion MNIST diffractive network and our results are summarized in Fig. S16, demonstrating that, for example, the addition of a 6th layer (learnable) to an already trained and fixed $D^2NN$ with N=5 improves its inference performance, performing slightly better than the performance of a $D^2NN$ with N=6 layers that were simultaneously trained. Also see Fig. S2 for an implementation of the same concept for MNIST: using a patch of 2 layers added to an existing/fixed $D^2NN$ (N=5), we improved our MNIST classification accuracy to 93.39%. The advantage of this Lego-like transfer learning or patching approach is that already fabricated and printed $D^2NN$ designs can be improved in performance by adding additional printed layers to them or replacing some of the existing diffractive layers with newly trained ones. This can also help us with the training process of very large network designs (e.g., N ≥ 25) by training them in patches, making it more tractable with state of the art computers."

## 5. Nomenclature

In their Comment, Wei *et al.* (*3*) claim that our $D^2NN$ framework should be termed as a computer generated volumetric hologram. Computer generated or not, a "volume hologram" in optics community is in general known as a "thick" hologram, and is mostly created by using a light source to expose a thick volume of a photosensitive material, creating 3D refractive index changes within its volume. That is why, we do not find the use of "volume hologram" a proper term for $D^2NN$ framework as it will be confusing for optics community.

An alternative terminology used in optics that could be related to $D^2NN$ framework is Diffractive Optical Elements (DOEs) (*4*, *5*). In our original nomenclature, the "D" of $D^2NN$ emphasizes on the word "Diffractive", and brings deep learning and diffractive optical surfaces into the same framework for implementation of optical machine learning and deep learning-based optical component design.

## 6. $L^2$ norm analysis

Equation (1) and the related analysis reported in the Comment by Wei *et al.* (*3*) are not in conflict with any of the results, analysis and conclusions of our manuscript *(1)* and therefore do not provide an amendment to the original teachings of Ref. *(1)*.

In the conclusion of their simple analysis, Wei *et al.* write: "*When two different images are in close similarity that the $L^2$ distance $\|\psi_0 - \phi_0\|_2$ is small and below a certain noise level, the system of Lin et al.'s will have a hard time to tell $\psi_0$ and $\phi_0$ apart, no matter how obviously different they are to human eyes…*" (*3*). Although it is not within the scope of our original submission *(1)*, one should emphasize that adversarial attacks on standard electronic deep neural networks with various nonlinear activation functions, can make them fail with almost a negligible $L^2$ distance between two images, no matter how obviously similar they are to the human eye. As detailed in Refs. (*6*–*8*), minimal $L^2$ norm perturbations, which are imperceptible for a human observer, can cause misclassification in state-of-the-art deep neural networks. Taking it further, Ref. (*9*) shows that such perturbations do not even need to be designed through complex algorithms or exhaustive search of parameters.

To summarize, in their simple analysis, Wei *et al.* merely report a truism, without a correction to the results or research methodology of Ref. *(1)*. Furthermore, their arguments about "human eye, $L^2$ distances *etc.*" are trivial conclusions *without* an amendment to or an improvement of our original teachings and conclusions of Ref. *(1)*.

## 7. Conclusion

In summary, the Comment by Wei *et al.* is regrettably ignorant of several important parts of our manuscript *(1)* related to optical nonlinearities and reconfigurability of $D^2NN$ framework. Since this unfortunate obliviousness is combined with a misunderstanding of the "depth feature" of an optical D$^2$NN, one can conclude that their 'mischaracterization' claim is entirely unjustified and, at best a confusion in nomenclature. All the results, the core conclusions and the methodology of research in our original manuscript remain entirely valid.

*References*

1. X. Lin *et al.*, All-optical machine learning using diffractive deep neural networks. *Science*. **361**, 1004–1008 (2018).

2. D. Mengu, Y. Luo, Y. Rivenson, A. Ozcan, Analysis of diffractive optical neural networks and their integration with electronic neural networks. *ArXiv181001916 Cs Physics* (2018) (available at http://arxiv.org/abs/1810.01916).

3. H. Wei, G. Huang, X. Wei, Y. Sun, H. Wang, Comment on All-optical machine learning using diffractive deep neural networks. *ArXiv180908360v1* (2018) (available at http://arxiv.org/abs/1809.08360v1).

4. O. K. Ersoy, *Diffraction, fourier optics, and imaging* (Wiley-Interscience, Hoboken, N.J, 2007).

5. J. W. Goodman, *Introduction to Fourier Optics* (Roberts and Company Publishers, 2005).

6. S. Gu, L. Rigazio, Towards Deep Neural Network Architectures Robust to Adversarial Examples. *ArXiv14125068 Cs* (2014) (available at http://arxiv.org/abs/1412.5068).

7. S.-M. Moosavi-Dezfooli, A. Fawzi, P. Frossard, in *2016 IEEE Conference on Computer Vision and Pattern Recognition (CVPR)* (IEEE, Las Vegas, NV, USA, 2016; http://ieeexplore.ieee.org/document/7780651/), pp. 2574–2582.

8. N. Papernot, P. McDaniel, X. Wu, S. Jha, A. Swami, in *2016 IEEE Symposium on Security and Privacy (SP)* (2016), pp. 582–597.

9. A. Kurakin, I. Goodfellow, S. Bengio, Adversarial examples in the physical world. *ArXiv160702533 Cs Stat* (2016) (available at http://arxiv.org/abs/1607.02533).